\documentclass[lettersize,journal]{IEEEtran}
\usepackage{amsmath,amsfonts,amssymb}
\usepackage{algorithmic}
\usepackage{algorithm}
\usepackage{array}
\usepackage[caption=false,font=normalsize,labelfont=sf,textfont=sf]{subfig}
\usepackage{textcomp}
\usepackage{stfloats}
\usepackage{url}
\usepackage{graphicx}
\usepackage{cite}
\usepackage{booktabs}
\usepackage{nicefrac}
\usepackage{float}
\usepackage{microtype}
\usepackage{xcolor}
\usepackage{colortbl}
\usepackage{multirow}
\usepackage{hyperref}
\hypersetup{
  pdftitle={Clay-CNN Hybrids: Leveraging Geo-Foundation Models as Auxiliary Context for Landslide Detection},
  pdfauthor={Binh Huong Vu},
  pdfkeywords={Geo-Foundation Models, Landslide Detection, Clay, Vision Transformers, U-Net, Remote Sensing},
  colorlinks=true,
  linkcolor=black,
  citecolor=black,
  urlcolor=blue
} 

\begin{document}

\title{Clay-CNN Hybrids: Leveraging Geospatial Foundation Models as Auxiliary Context for Landslide Detection}

\author{Binh~Huong~Vu%
\thanks{B.~Vu was with Harvard University, Cambridge, MA 02138 USA
(e-mail: bvu@college.harvard.edu).}%
}

\markboth{}%
{Vu: Clay-CNN Hybrids for Landslide Detection}

\maketitle

\begin{abstract}
Rapid post-event landslide mapping is essential for disaster response but remains difficult
to automate due to extreme class imbalance. This study evaluates whether Clay v1.5, a Geospatial Foundation Model (GFM), can
improve pixel-level landslide segmentation on the Landslide4Sense (L4S) benchmark (3,799
training chips, 14 Sentinel-2 and terrain bands, $\sim$2\% positive pixels). We compare three
strategies: Clay as the primary encoder with multi-scale residual terrain fusion, a U-Net
backbone augmented with Clay semantic context at the bottleneck, and a standard U-Net
baseline. The hybrid U-Net + Clay model with two-stage Low-Rank Adaptation (LoRA) achieved the best test F1 of
64.5$\pm$1.8\% over three seeds, surpassing the Clay-only backbone (55.2 $\pm$ 3.6\%) and the U-Net baseline (59.9\%). Clay
as a standalone encoder underperformed the U-Net due to the absence of multi-scale skip
connections, but its pretrained representations consistently improved performance when
injected as auxiliary context. These findings suggest that GFMs are
most effective for landslide detection when they complement spatially-detailed
convolutional architectures rather than replace them.
\end{abstract}


\section{Introduction}

\IEEEPARstart{L}{andslides} rank among the most pervasive geohazards worldwide, causing major human casualties and extensive
environmental and economic damage \cite{ref1}. The complexity of assessing these risks is particularly acute in seismically active regions, where earthquake-induced landslides can dramatically escalate threats to infrastructure and human life \cite{ref22, ref23}.
Furthermore, climate change is projected to increase landslide incidence globally, disproportionately affecting regions in Southeast Asia and Latin America \cite{ref24}. These areas often lack the systematic landslide inventories required to train traditional supervised detection models, creating a critical ``data-scarce'' regime \cite{ref2}.

Although advances have been made in susceptibility mapping and early warning systems, rapid post-event mapping remains a significant challenge. Delineating new landslide scars from satellite imagery within hours of an event is essential for emergency response and for generating the ground truth data needed to refine future warning systems. Automated detection through satellite-based semantic segmentation offers a scalable solution, but is hindered by two primary technical obstacles.

\begin{enumerate}
  \item \textit{Spectral Similarity:} Landslide scars often appear spectrally similar to bare soil \cite{ref18}.
  \item \textit{Extreme Class Imbalance:} In benchmarks such as Landslide4Sense, landslide pixels can comprise as little as 2\% of all pixels \cite{ref27}.
\end{enumerate}

Current deep learning approaches, predominantly based on U-Net architectures, rely heavily on task-specific training data---a major constraint during active disasters when labeled examples of new events are unavailable \cite{ref3, ref4}. In India, Sreekumar \textit{et al.}\ (2025) used multiscale and attention-based U-Net
architectures trained on pre- and post-event imagery, incorporating auxiliary terrain variables derived from a Digital Elevation Model (DEM) and slope gradient to capture geomorphic controls on slope failure \cite{ref3}. In a demonstration of operational potential, Nava \textit{et al.}\ (2025) showed that fine-tuning pretrained segmentation models on event-specific data following the 2024 Hualien earthquake in Taiwan enabled the identification of over 7,000 landslide polygons covering 75~km\textsuperscript{2} within
approximately three hours of image acquisition \cite{ref4}. These studies highlight both the promise of deep learning for rapid mapping and its continued reliance on task-specific training data, a significant constraint in disaster response contexts where labeled examples of new events are unavailable at the moment they are most needed.

Vision foundation models offer a potential route around this constraint by providing generalizable representations
pretrained on large, diverse image collections. Yang \textit{et al.}\ (2025) \cite{ref5} applied the Segment Anything Model (SAM) \cite{ref29} to
landslide identification by fusing an RGB branch with a DEM branch, achieving improved F1 over conventional Mask Region-Based Convolutional Neural Network
(Mask R-CNN) baselines. More directly relevant to this study, emerging work has begun to evaluate Geospatial Foundation Models
(GFMs)---pretrained on large corpora of multispectral satellite imagery---for dense segmentation tasks.
Szwarcman \textit{et al.}\ (2026) applied Prithvi-EO-2.0 to landslide segmentation on L4S, achieving F1 of 60.7\%
versus a U-Net baseline of 59.7\% \cite{ref6}. Kaushik \textit{et al.}\ (2026) subsequently benchmarked three GFMs
across flood inundation tasks and found Clay v1.5 \cite{ref12} to consistently outperform Prithvi-EO-2.0 across
data-scarce regimes and multiple sensor types, while identifying CNN late-fusion as a promising future
direction \cite{ref7}. While GFMs like Clay v1.5 \cite{ref12} offer a potential solution by providing generalizable representations pretrained on massive, diverse datasets, there has been no systematic evaluation of how these models can be best integrated with convolutional architectures to optimize landslide detection. This study addresses that gap by investigating how GFM-CNN hybrid architectures can leverage both the spectral generalization of foundation models and the spatial precision of traditional networks.

This paper investigates how GFMs may improve semantic segmentation of landslides in scarce data regimes. To address this, we make the following contributions:

\begin{itemize}
  \item We introduce two novel architectures: in the first (Arch~1), Clay is used as the primary encoder with multi-scale residual terrain fusion; in the second (Arch~2), Clay serves as auxiliary bottleneck context within a U-Net backbone.
  \item We perform systematic ablation over encoder freezing, LoRA fine-tuning, two-stage training, loss formulation, and
        terrain integration.
  \item We incorporate Monte Carlo Dropout (MC Dropout) uncertainty quantification \cite{ref8} and Gradient-weighted Class Activation Mapping (Grad-CAM) interpretability analysis \cite{ref9} to diagnose model
        behavior beyond aggregate metrics.
\end{itemize}

The remainder of this paper is organized as follows. Section~\ref{sec:data} describes the dataset, preprocessing steps, and evaluation protocol. Section~\ref{sec:methods} details the three architectures and key design decisions. Section~\ref{sec:results} presents experimental results and ablation analysis. Section~\ref{sec:discussion} discusses implications for the use of foundation
models in geohazard mapping. Section~\ref{sec:conclusion} concludes with limitations and directions for future work.

\section{Dataset and Preprocessing}
\label{sec:data}

The study utilizes the Landslide4Sense (L4S) benchmark, originally from the 2022 IEEE JSTARS competition, accessed via HuggingFace,
the only publicly available version that includes test set ground truth masks \cite{ref10,ref27}. The dataset consists of 3,799 training, 245 validation, and 800 test chips, each formatted as a $128 \times 128$ pixel image (chips) at 10~m resolution.

\begin{figure}[!t]
  \centering
  \includegraphics[width=3.2in]{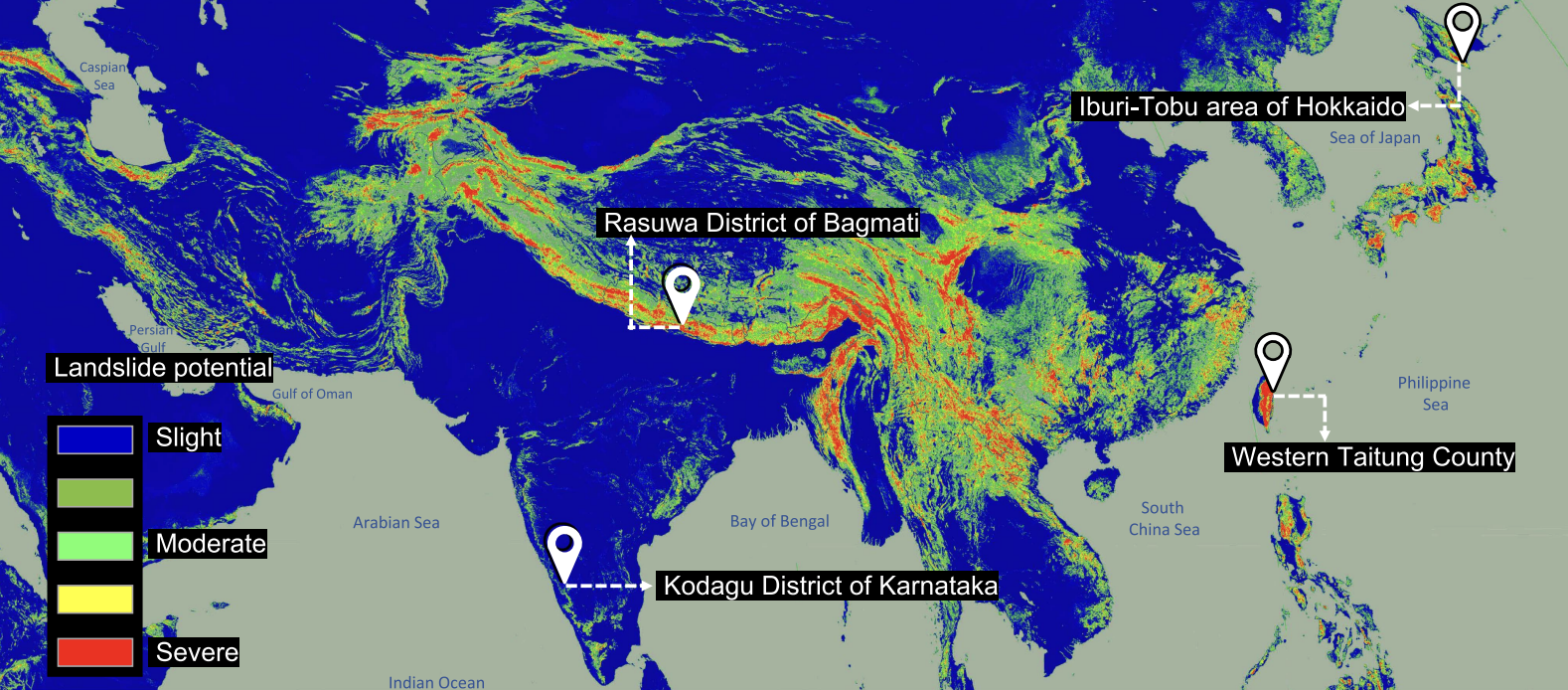}
  \caption{Geographic distribution of the Landslide4Sense training data across landslide-prone regions of Asia, overlaid on a landslide susceptibility map (blue = slight potential, red = severe). Labeled markers indicate the four training locations: Rasuwa District of Bagmati (Nepal), Iburi-Tobu area of Hokkaido (Japan), Kodagu District of Karnataka (India), and Western Taitung County (Taiwan). Adapted from the 2022 IEEE JSTARS competition \cite{ref30}.}
  \label{fig:dataset}
\end{figure}

To ensure the model generalizes to new events, the dataset employs a geographically stratified ``leave-out-location'' strategy. This approach prevents spatial leakage by ensuring that data from specific landslide events---such as those in the Rasuwa District of Bagmati, the Iburi-Tobu area of Hokkaido, the Kodagu District of Karnataka, and Western Taitung County---are not split across the training and testing sets (Fig.~\ref{fig:dataset}) \cite{ref30}.

Each data chip contains 14 input channels providing a combination of optical and geomorphic information:

\begin{itemize}
  \item \textbf{Sentinel-2 (Bands 1--12):} Covers the spectral range from coastal aerosol through short-wave infrared (SWIR).
  \item \textbf{Topographic data (Bands 13--14):} Includes a digital elevation model (DEM) and derived slope from ALOS PALSAR (acquired between 2006--2019), resampled from its native 12.5\,m resolution to match the 10\,m Sentinel-2 grid.
\end{itemize}

\begin{figure*}[!t]
  \centering
  \includegraphics[width=0.90\textwidth]{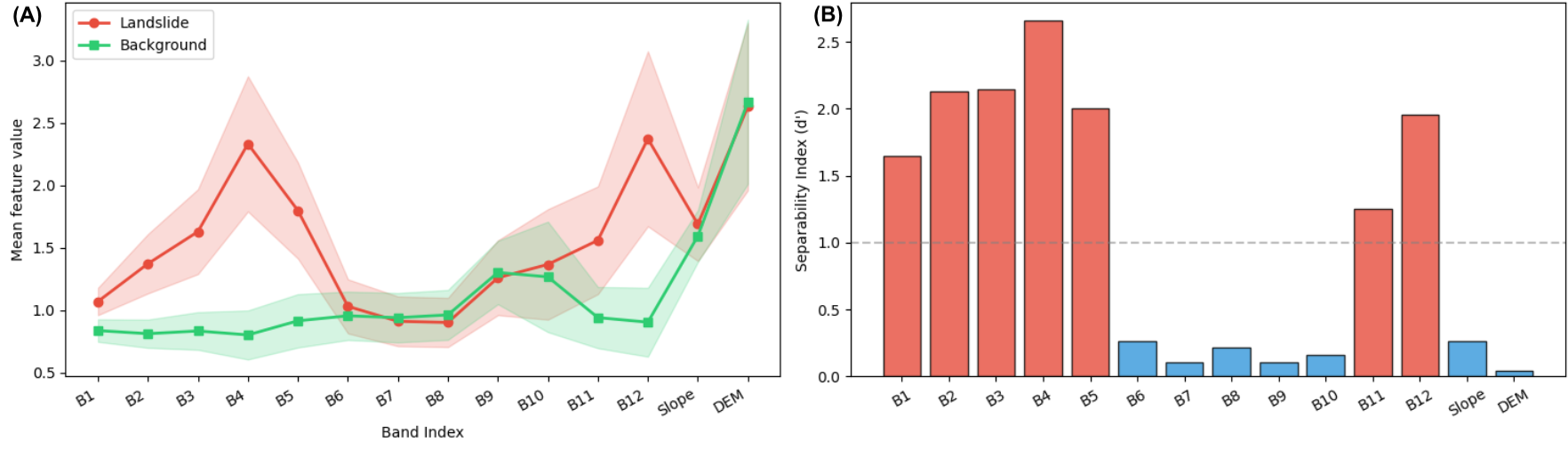}
  \caption{Exploratory data analysis. (A) Mean spectral signatures of landslide vs.\ background pixels across the 14 input bands. (B) Per-band separability index $d'$; red bars mark highly discriminative bands ($d' > 1.0$), concentrated in the visible, red-edge, and SWIR regions.}
  \label{fig:eda}
\end{figure*}

The dataset exhibits extreme class imbalance: landslide pixels comprise only about 2\% of the training labels. Most chips contain very few positive pixels, motivating the use of weighted loss functions during training. A per-band separability analysis using the separability index ($d^{\prime}$) shows that different spectral regions contribute uneven signal (Fig.~\ref{fig:eda}):

\begin{itemize}
  \item \textbf{High discriminability ($d^{\prime} > 1.0$):} Visible (B1--B4), red-edge (B5--B7, B8A), and SWIR (B11--B12) bands are most informative for identifying landslide scars.
  \item \textbf{Low discriminability ($d^{\prime} < 0.5$):} Near-infrared (NIR) and water vapor bands carry substantially less signal.
  \item \textbf{Geomorphic context:} Terrain features (DEM and slope) show limited pixel-level separability, but could provide physically meaningful context for identifying areas where failures are plausible.
\end{itemize}

Accordingly, terrain features are treated as auxiliary inputs in Arch~1. Their low pixel-level discriminability motivates the use of a learnable scalar $\alpha$, which allows the model to control how much geomorphic context to incorporate \cite{ref3}.

To prepare the data for the Clay v1.5 foundation model, several specialized steps were taken:

\begin{itemize}
  \item \textbf{Normalization:} Statistics (mean and standard deviation) were computed solely from the training split using streaming accumulation to prevent data leakage.
  \item \textbf{Wavelength-Aware Embedding:} For the Clay encoder, per-band physical wavelengths were registered in a custom metadata YAML. This allows the model's wavelength-aware patch embedding to correctly associate each channel with its specific spectral position.
  \item \textbf{Invalid value handling:} NaN/Inf pixels are replaced with the channel training mean before normalization,
        avoiding zero-fill artifacts that would distort the normalized distribution.
  \item \textbf{Geometric augmentation (training only):} Random horizontal/vertical flip and 90$^\circ$ rotation, applied
        identically to image and mask.
\end{itemize}

\section{Methods}
\label{sec:methods}

All experiments build on Clay v1.5, a geospatial Vision Transformer pretrained via masked autoencoding on over 70
million Earth observation patches spanning diverse sensors, seasons, and global locations \cite{ref12}. Rather than expecting
fixed input channels, Clay projects each band through a wavelength-aware embedding mapping physical
wavelengths to a continuous embedding space, enabling flexible integration with any band combination without
retraining. This sensor-agnostic design makes it well-suited for scarce-label settings like L4S, where generalizable
pretrained representations reduce dependence on large task-specific datasets.

\subsection{Baseline: Competition U-Net}

A standard U-Net with four encoder stages and a bottleneck (channel widths 64, 128, 256, 512, 512), MaxPool2d downsampling,
bilinear upsampling decoder with full skip connections at all four scales, and BatchNorm throughout, is used as the baseline model. Each encoder
and decoder stage uses a DoubleConv block consisting of two Conv$3{\times}3$, BatchNorm, ReLU sequences. All 14 bands
are fed as a single input tensor. The output is a 2-class softmax trained with cross-entropy loss and Adam
($\text{LR}{=}1{\times}10^{-3}$, $\text{WD}{=}5{\times}10^{-4}$) for 5,000 epochs. This design follows the original competition
implementation closely \cite{ref11}, serving as the CNN-only reference point to isolate the effect of the encoder across
subsequent architectures.

\subsection{Architecture 1: Clay Backbone with Multi-Scale Terrain Fusion}

\begin{figure*}[!t]
  \centering
  {\includegraphics[width=2.0in]{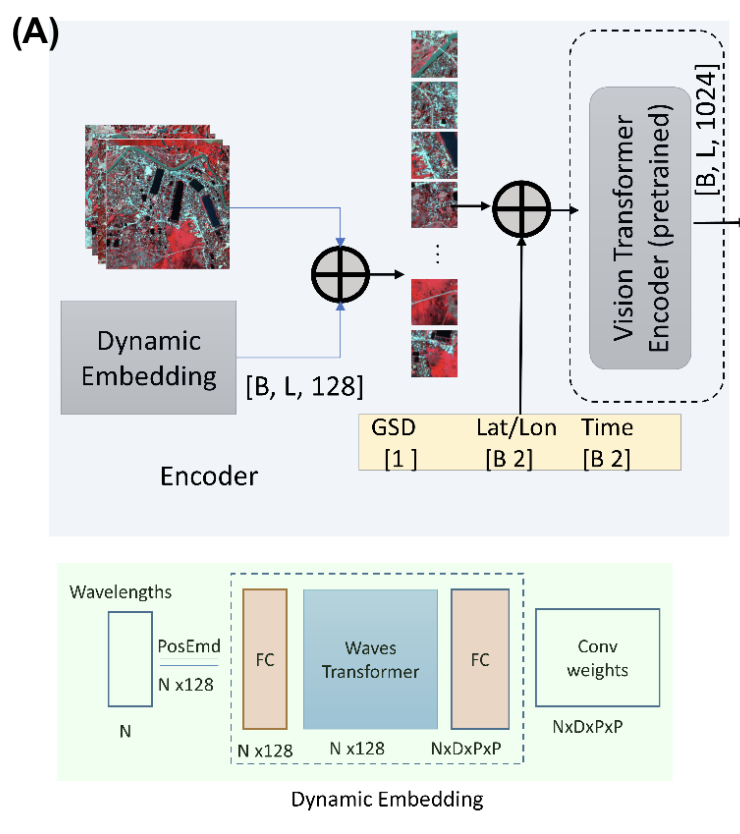}%
  }
  \hfil
  {\includegraphics[width=4.7in]{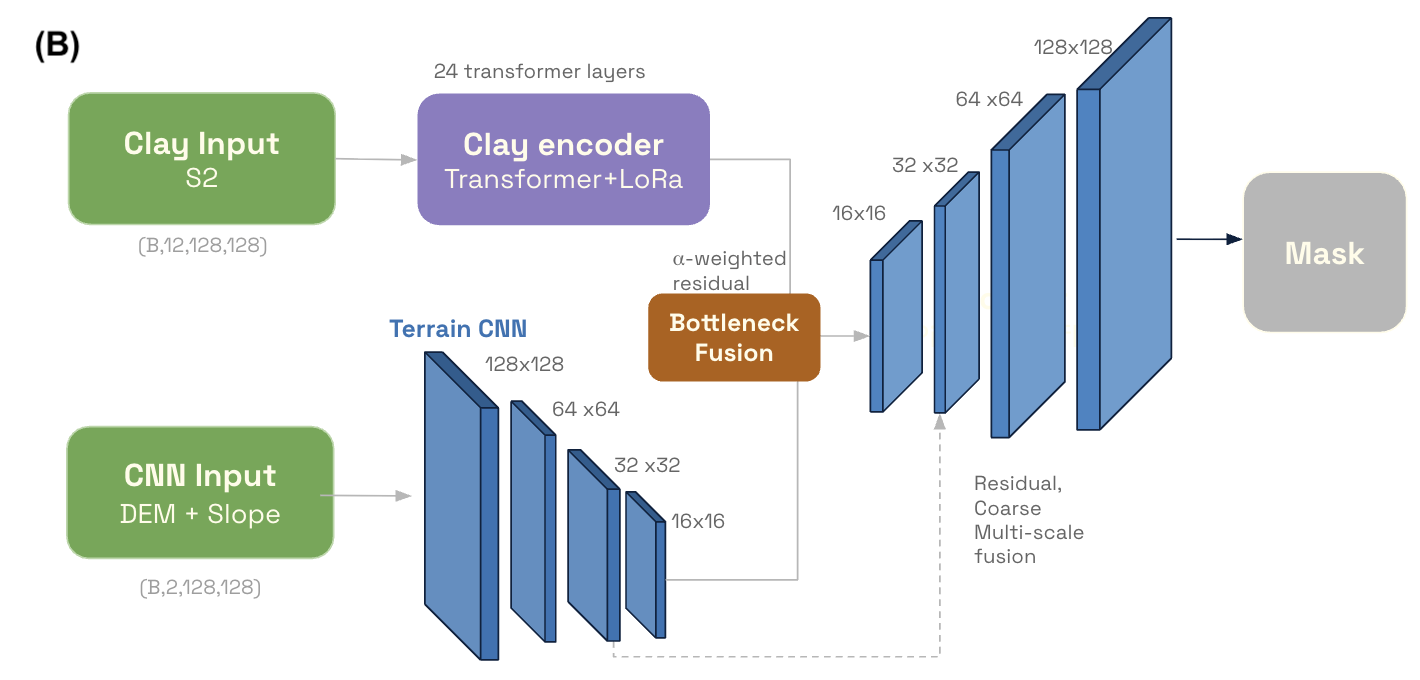}%
  }
  \caption{Architecture 1: Clay as primary encoder with multi-scale terrain fusion. (A) The Clay v1.5 ViT encoder design from Kaushik \textit{et al.}\ \cite{ref7}. (B) Proposed architecture: 12 Sentinel-2 bands are processed by the frozen/LoRA-adapted Clay encoder into a single $16{\times}16$ feature map, while DEM and slope pass through a separate convolutional TerrainPyramid. Terrain features are fused residually at the two coarsest decoder scales via a learnable scalar $\alpha$ initialized to zero.}
  \label{fig:arch1_full}
\end{figure*}

Clay serves as the primary encoder, taking 12 Sentinel-2 bands packaged as a datacube with physical wavelengths,
constant time and geolocation encodings (set to 1), and Ground Sample Distance (GSD)$=$10\,m \cite{ref7}. Terrain channels are
excluded since DEM and slope are not
spectral measurements, and feeding them through Clay's wavelength-aware patch embedding would be semantically
inconsistent with its pretraining. The ViT tokenizes the $128{\times}128$ input with patch size 8, producing 256 patch tokens
projected to dimension 1024, processed through 24 transformer blocks of self-attention and MLP with residual
connections, and reshaped to a spatial feature map of shape $(B, 1024, 16, 16)$. This is projected to 96 channels via a
$1{\times}1$ convolution, BatchNorm, and ReLU.

DEM and slope are processed through a separate TerrainPyramid consisting of four ConvBNReLU layers with
stride-2 downsampling. The decoder uses three stages of bilinear upsampling (16, 32, 64, then 128). Terrain features are fused at the
bottleneck ($16{\times}16$) and first decoder stage ($32{\times}32$) via a ResidualTerrainFusionBlock:

\begin{equation}
  \text{output} = f\!\left(\,\text{proj}(x_{\text{dec}}) + \alpha \cdot g(x_{\text{terrain}})\right)
  \label{eq:terrain_fusion}
\end{equation}

\noindent where $\alpha$ is a learnable scalar initialized to 0.0, meaning terrain starts with zero influence and the model learns
how much geomorphic context to incorporate. Fusion is restricted to coarse scales only ($32{\times}32$ and $16{\times}16$ resolution),
avoiding injection of noisy high-resolution terrain into fine-grained predictions. The segmentation head consists of
Conv$3{\times}3$, BatchNorm, ReLU, Dropout (0.1), and Conv$1{\times}1$. All new layers outside the pretrained Clay encoder---including the decoder, TerrainPyramid,
fusion blocks, and projection layers---are initialized with He (Kaiming) initialization \cite{ref13}. Architecture~1 has no
encoder skip connections; all spatial detail must be reconstructed from the single $16{\times}16$ Clay output, a fundamental
structural limitation relative to the U-Net.

\subsection{Architecture 2: Hybrid U-Net + Clay}

\begin{figure*}[!t]
  \centering
  \includegraphics[width=0.75\textwidth]{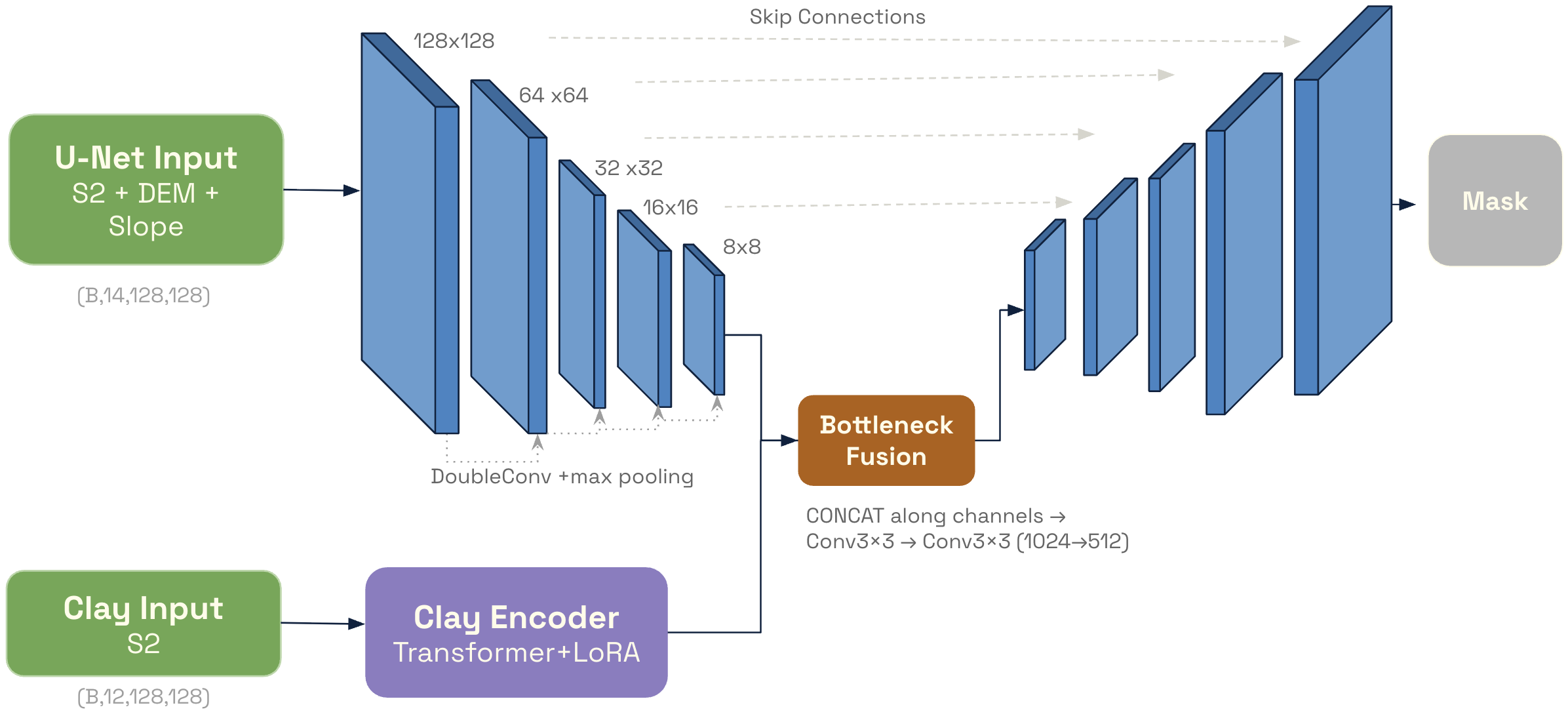}
  \caption{Architecture 2: Hybrid U-Net + Clay. The U-Net backbone encodes all 14 bands through four downsampling stages with full skip connections, while Clay independently processes the 12 Sentinel-2 optical bands. Clay's output is projected, downsampled to $8{\times}8$, and fused with the U-Net bottleneck via channel concatenation followed by two Conv$3{\times}3$ blocks. The fused representation is decoded with standard U-Net skip connections to produce the segmentation mask.}
  \label{fig:arch2}
\end{figure*}

The U-Net provides the primary spatial backbone, taking all 14 bands and encoding them through four
downsampling stages with channel widths 64, 128, 256, 512, and 512 via MaxPool2d to an $8{\times}8$ bottleneck. Full skip
connections are preserved at all four spatial scales. Clay independently processes the 12 Sentinel-2 optical bands.
Clay outputs a feature map of shape $(B, 1024, 16, 16)$, which is projected to 512 channels
via a $1{\times}1$ convolution, BatchNorm, and ReLU, then spatially aligned to the U-Net bottleneck
resolution via bilinear downsampling ($16{\times}16 \rightarrow 8{\times}8$). This mild $2{\times}$
reduction is necessary to match the four-stage U-Net encoder's spatial output and is not expected
to cause significant information loss given the coarse semantic nature of ViT bottleneck features.

Fusion is bottleneck-only: the U-Net bottleneck and the projected Clay map are concatenated along the channel
dimension to produce a 1024-channel tensor, which is compressed back to 512 channels via two Conv$3{\times}3$,
BatchNorm, ReLU blocks. This fused representation enters the four-stage U-Net decoder, where each stage
bilinearly upsamples the features, concatenates with the matching encoder skip connection, and applies a
DoubleConv block followed by Dropout2d (10\%). Clay contributes at a single spatial scale only---one global feature
map---while the U-Net supplies the full multi-scale spatial hierarchy and boundary detail through its skip
connections.

\subsection{Training and Optimization}

All three architectures share the same core training configuration where applicable:

\begin{itemize}
  \item \textbf{Loss function.} The primary loss combines Binary Cross-Entropy (BCE) and Lov\'{a}sz hinge loss \cite{ref14}:
        \begin{equation}
          \mathcal{L} = 0.9 \times \text{BCE}(\text{pos\_weight}{=}25) + 0.1 \times \mathcal{L}_{\text{Lov\'{a}sz}}
          \label{eq:loss}
        \end{equation}
      The positive weight of 25 counteracts the 49:1 class imbalance. Lov\'{a}sz provides a differentiable
surrogate for the Jaccard index, directly optimizing toward F1 rather than pixel-independent accuracy,
with the 0.9/0.1 split preserving BCE dominance for stable early gradients.

  \item \textbf{Optimizer and scheduler.} AdamW with weight decay $1{\times}10^{-3}$. Learning rate follows cosine
        annealing with $\eta_{\min} = 1{\times}10^{-6}$. Separate learning rate groups are maintained for the
        decoder/fusion layers and LoRA adapters.

  \item \textbf{Two-stage training (Architectures 1 and 2).} \cite{ref15} Stage~1 freezes the Clay encoder entirely and trains only the
        decoder, terrain branch, fusion layers, and segmentation head at LR$\,{=}\,3{\times}10^{-4}$. This stabilizes the
        decoder before encoder adaptation begins. At epoch 11, Stage~2 inserts LoRA adapters into all 24 Clay
        transformer blocks (targeting attention Q/K/V projections and MLP layers, rank$\,{=}\,$8, $\alpha{=}$8) and trains them at a lower LR$\,{=}\,1{\times}10^{-5}$ for Arch~1 and $3{\times}10^{-5}$ for Arch~2, reflecting Arch~2's bottleneck fusion which dilutes the LoRA gradient signal. The decoder LR decreases to $1{\times}10^{-4}$ for both. Switching at epoch 11, when Stage~1 begins to plateau, ensures the decoder is functional but still able to co-adapt with LoRA.

  \item \textbf{Regularization.} Dropout2d at 10\% in all decoder and fusion layers. Gradient clipping.

  \item \textbf{Early stopping.} Monitored on validation F1 with patience of 10 epochs. Early stopping is only
        active after Stage~2 begins to give each stage a full runway.

  \item \textbf{Batch size.} 8.
\end{itemize}

\subsection{Evaluation Protocol}

All models are evaluated on pixel-wise F1 score, matching the original competition metric \cite{ref16}:

\begin{equation}
  \text{F1} = \frac{2 \times \text{Precision} \times \text{Recall}}{\text{Precision} + \text{Recall}}
  \label{eq:f1}
\end{equation}

Converting model output probabilities to binary predictions requires a classification threshold. Under severe class
imbalance, the optimal decision boundary rarely falls at 0.5. Since threshold selection occurs post-training on
fixed model weights, it does not influence parameter updates. A sweep over candidate thresholds on the validation
set selects the value maximizing validation F1, which is then locked and applied to the test set exactly once,
preventing test set contamination.

\section{Results}
\label{sec:results}

Quantitative results across all models are reported in Table~\ref{tab:ablation}. Interpretability analysis is conducted on
Architecture~2 (best model).

\begin{table*}[!t]
\caption{Ablation Results on the Landslide4Sense Test Set. Best F1 Shown in Bold.}
\label{tab:ablation}
\centering
\setlength{\tabcolsep}{6pt}
\begin{tabular}{@{}clcccc@{}}
\toprule
\# & Model & Precision (\%) & Recall (\%) & F1 (\%) & Threshold \\
\midrule
\multicolumn{6}{@{}l}{\textit{Baseline}} \\
1 & U-Net (competition) & 52.5 & 69.9 & 59.9 & 0.50 \\
\midrule
\multicolumn{6}{@{}l}{\textit{Architecture 1: Clay Backbone}} \\
2 & Clay (frozen) + terrain concat & 39.6 & 50.8 & 49.3 & 0.50 \\
3 & Clay (LoRA) + terrain concat & 29.0 & 33.6 & 30.2 & 0.50 \\
4 & Clay (LoRA) + multiscale terrain fusion + tuned loss & 46.9 & 57.2 & 54.2 & 0.50 \\
5a & Clay (LoRA) + multiscale terrain fusion + two-stage & 51.5 $\pm$ 2.9 & 59.6 $\pm$ 4.6 & 55.2 $\pm$ 3.6 & 0.60 \\
5b & \quad $\hookrightarrow$ same model, fixed $t=0.50$ & 44.7 $\pm$ 4.6 & 68.2 $\pm$ 12.7 & 53.4 $\pm$ 2.6 & 0.50 \\
\midrule
\multicolumn{6}{@{}l}{\textit{Architecture 2: Hybrid U-Net + Clay}} \\
6 & U-Net + Clay (frozen) & 60.3 & 63.2 & 61.7 & 0.70 \\
7a & U-Net + Clay (LoRA) + two-stage & 62.2 $\pm$ 3.0 & 67.1 $\pm$ 0.6 & \textbf{64.5 $\pm$ 1.8} & 0.78 \\
7b & \quad $\hookrightarrow$ same model, fixed $t=0.50$ & 40.0 $\pm$ 0.7 & 87.2 $\pm$ 2.2 & 54.8 $\pm$ 0.7 & 0.50 \\
\bottomrule
\multicolumn{6}{@{}p{6in}@{}}{\footnotesize Rows with $\pm$ report mean $\pm$ standard deviation across $n=3$ seeds; other rows are single runs. Model~1 is the published L4S competition baseline \cite{ref30}. Rows 5a/7a use per-seed validation-selected thresholds; 5b/7b evaluate the same models at fixed $t=0.50$.}
\end{tabular}
\end{table*}

\textbf{Ablation Results:} Table~\ref{tab:ablation} presents the full ablation progression. Clay as a frozen standalone encoder (Model~2, F1=49.3\%) underperforms the U-Net baseline (Model~1, F1=59.9\%), confirming that the ViT's single-scale output without skip connections is insufficient for precise pixel-level segmentation. Applying LoRA without a stable decoder (Model~3, F1=30.2\%) actively degrades performance. Multiscale terrain fusion, loss tuning, and two-stage training progressively recover performance to a multi-seed mean of 55.2 $\pm$ 3.6\% (Model~5a), with the two-stage strategy contributing +1.0~pp over Model~4 by stabilizing the decoder before encoder adaptation. This still falls below the U-Net baseline, revealing that Clay's lack of encoder skip connections is a structural limitation fine-tuning alone cannot overcome. Architecture~2 resolves this: injecting a frozen Clay context in the bottleneck (Model~6, F1=61.7\%) already surpasses all Architecture~1 variants, and the two-stage LoRA achieves the best overall result with a multi-seed mean of 64.5 $\pm$ 1.8\% (Model~7a), a \textbf{gain of 4.6~pp} over the baseline.

\begin{figure}[!t]
  \centering
  \includegraphics[width=1.70in]{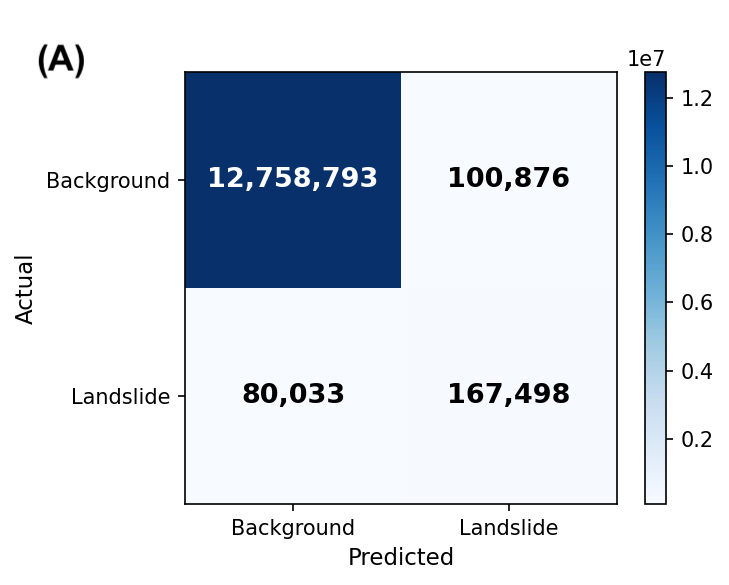}%
  \hfil
  \includegraphics[width=1.65in]{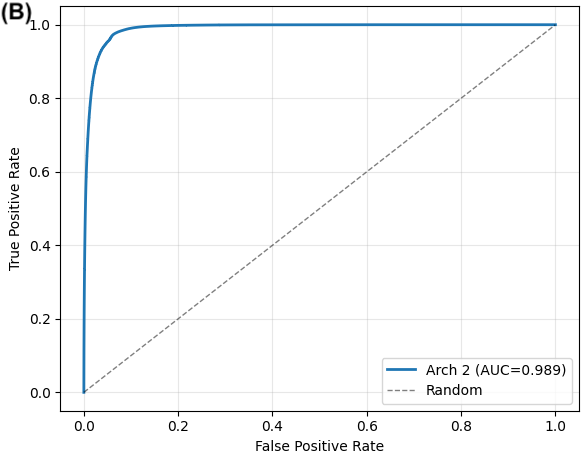}%
  \caption{Test-set performance of Architecture 2 (Model 7a, representative seed 42). (A) Pixel-level confusion matrix at the validation-selected threshold $t{=}0.80$, yielding F1 = 64.9\%, precision = 62.4\%, and recall = 67.7\%. (B) ROC curve with AUC = 0.989, indicating strong discriminative ranking; the optimal operating threshold lies well above the 0.5 default due to class imbalance.}
  \label{fig:arch2_performance}
\end{figure}

\textbf{Best Model Performance:} Fig.~\ref{fig:arch2_performance}(A) shows the pixel-level confusion matrix for a representative Architecture~2 run (seed=42, $t=0.80$): the model correctly identifies 167,498 of 247,531 true landslide pixels (recall=67.7\%) and falsely flags only 100,876 of 12,859,669 background pixels (FPR=0.78\%), yielding F1=64.93\% at precision~$62.41\%$. Compared to the U-Net baseline (P=52.5\%, R=69.9\%), Architecture~2 trades modest recall for substantially higher precision, with false negatives dominating the error mode, specifically for small or spectrally ambiguous patches rather than over-prediction. Multi-seed runs confirm stability: F1=64.5$\pm$1.8\% across $n=3$ seeds (Table~\ref{tab:ablation}), with recall variance especially tight ($67.1\pm0.6\%$).

Evaluating the same models at fixed $t=0.50$ collapses F1 to $54.8\pm0.7\%$ (Row~7b), a 9.7~pp drop driven by precision fall to $40.0\%$ as recall jumps to $87.2\%$. The high ROC-AUC = 0.989 (Fig.~\ref{fig:arch2_performance}(B)) confirms that the model has learned a strong discriminative ranking, but its imbalance-induced probability distribution shifts the optimal decision boundary well above the default 0.5. The validation-driven threshold sweep is therefore a necessary calibration step under severe class imbalance.

\begin{figure*}[!t]
  \centering
  \includegraphics[width=0.79\textwidth]{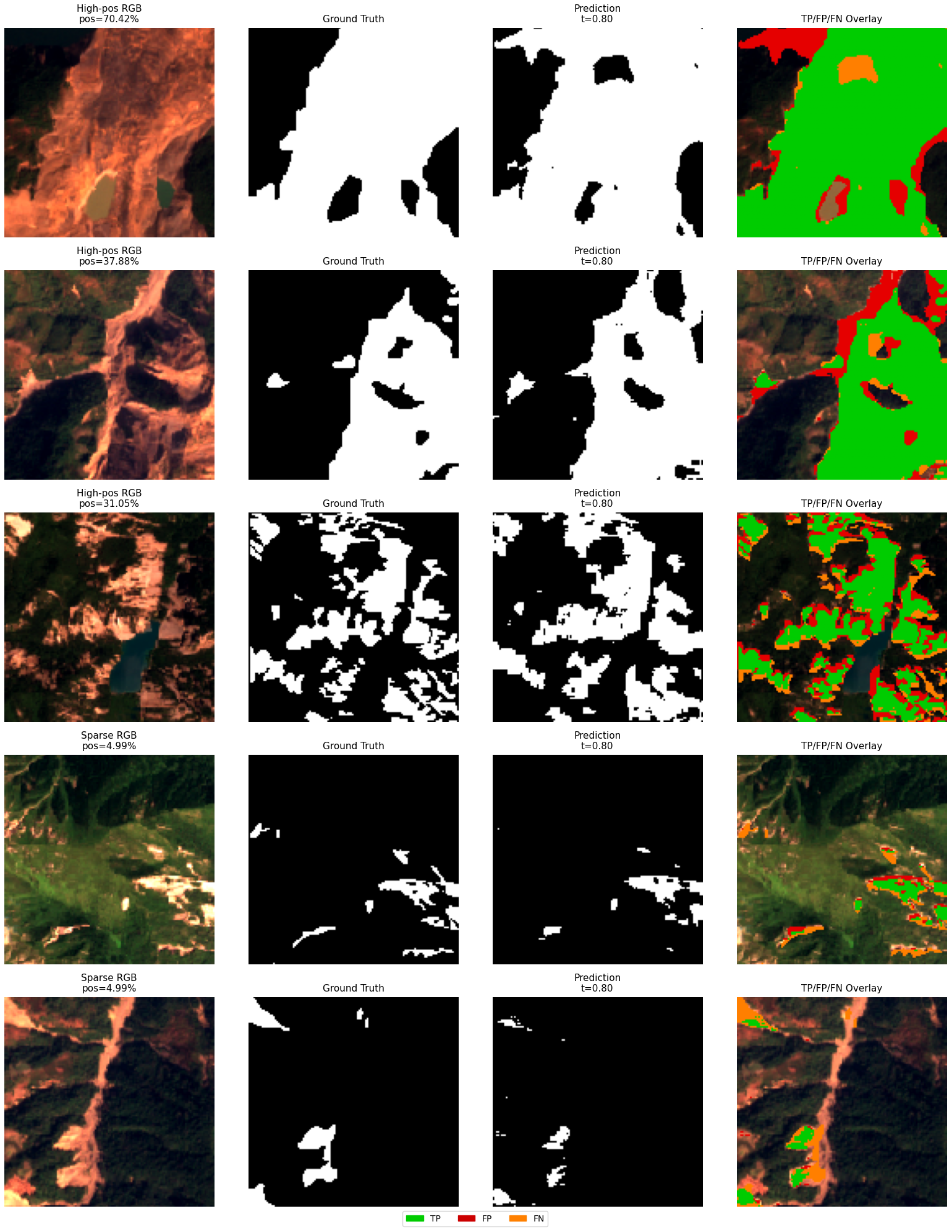}
  \caption{Qualitative predictions of Architecture 2  on six test chips ordered by decreasing positive pixel fraction (70.42\%--4.99\%; representative seed 42). Columns: Sentinel-2 RGB composite, ground truth mask, binary prediction at $t{=}0.80$, and error overlay (green = true positive, red = false positive, orange = false negative). Dense chips (rows 1-3) show accurate scar-level localization with errors confined to boundaries; sparse chips (bottom rows) exhibit missed small patches and fragmented predictions.}
  \label{fig:qualitative}
\end{figure*}

Fig.~\ref{fig:qualitative} shows predictions on six test chips spanning positive fractions of 4.99\%--70.42\%. For high-density
chips (pos $>$ 30\%), the model captures landslide extent well, with false negatives concentrating at scar
boundaries and in small detached fragments. False positives cluster in spectrally similar bare soil adjacent to true
scars. For sparse chips (pos $\approx$ 5\%), performance degrades noticeably---the model either misses small patches
entirely or produces fragmented predictions with poor spatial correspondence, consistent with weak gradient signal
from near-empty chips during training and limited spatial context at $128{\times}128$ resolution.

\textbf{Model Uncertainty Analysis:} MC Dropout uncertainty ($T{=}20$ forward passes) is compared between the frozen Clay baseline (Model~6) and the
best hybrid model (Model~7a) in Table~\ref{tab:mcdropout}.

\begin{table}[!t]
  \centering
  \caption{MC Dropout Uncertainty ($\sigma$) by Prediction Category for Frozen Clay (Model~6) and LoRA Two-Stage (Model~7a)}
  \label{tab:mcdropout}
  \begin{tabular}{lcc}
    \toprule
    & \textbf{Frozen Clay} & \textbf{LoRA Two-Stage} \\
    & \textbf{(Model 6)} & \textbf{(Model 7a)} \\
    \midrule
    Mean $\sigma$ (all) & 0.0133 &  0.00248 \\
    Mean $\sigma$ (TP)  & 0.0227 & 0.0351 \\
    Mean $\sigma$ (FP)  & 0.0270 &  0.0450 \\
    Mean $\sigma$ (FN)  & 0.0309 & 0.0431 \\
    Mean $\sigma$ (TN)  & 0.0130 & 0.00170 \\
    \bottomrule
  \end{tabular}
\end{table}

Two observations are notable. First, the uncertainty ordering follows TN $<$ TP $<$ FN $<$ FP in both models, indicating the model is most uncertain on its errors. Second, after Dirichlet calibration, Model~7a exhibits \textit{higher} uncertainty than the frozen baseline on every error category ($\sigma_{\text{FP}}$: $+67\%$, $\sigma_{\text{FN}}$: $+40\%$, $\sigma_{\text{TP}}$: $+55\%$), while remaining appropriately low on confidently correct backgrounds ($\sigma_{\text{TN}}=0.0017$) \cite{kull2019}. The lower aggregate mean ($\sigma=0.0025$) reflects this concentration of certainty on the 98\% of pixels that are easy negatives, not systematic overconfidence. The model is uncertain where it makes mistakes, which is the desired behavior for operational deployment.

Per-chip uncertainty maps show that
high-uncertainty regions concentrate along the boundaries of landslides and in sparse areas where the
model struggles the most, which is consistent with the MC Dropout analysis showing high mean $\sigma_{\text{FN}}=0.0431$ and $\sigma_{\text{FP}}=0.0450$. This spatial alignment between
uncertainty and error-prone regions is a desirable property, indicating that the model is aware
of where its predictions are unreliable.

\begin{figure*}[!t]
  \centering
  \includegraphics[width=0.85\textwidth]{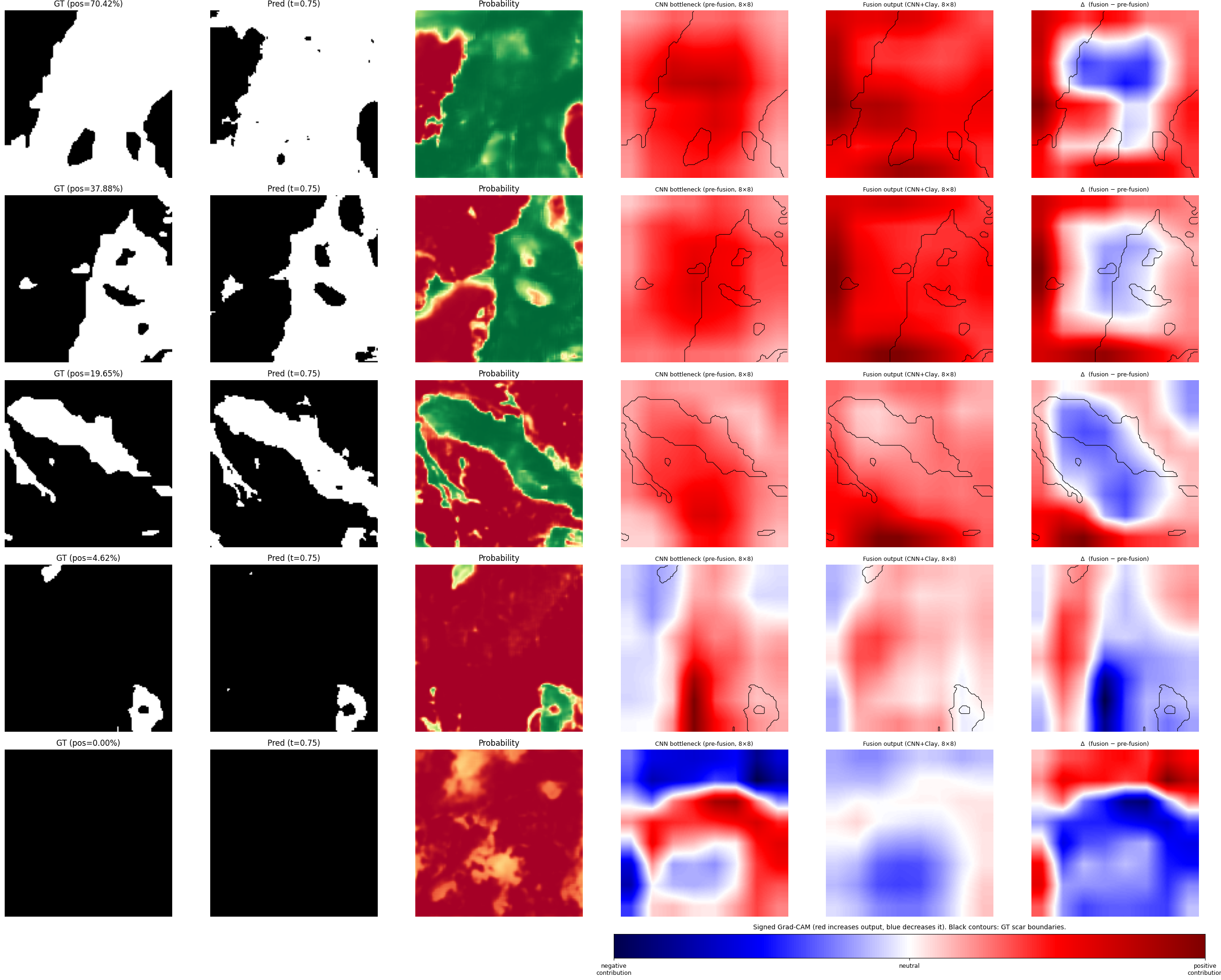}
  \caption{Signed Grad-CAM fusion analysis for Architecture 2 on five test chips spanning dense to empty scar coverage. Columns: ground truth mask, binary prediction, predicted probability, Grad-CAM at the CNN bottleneck (pre-fusion, $8{\times}8$), Grad-CAM at the fusion output (CNN+Clay, $8{\times}8$), and their difference $\delta$ = fusion $-$ pre-fusion. Red indicates positive contribution to the landslide class, blue negative; black contours mark ground-truth scar boundaries. Fusion systematically broadens positive contribution outside scar interiors, consistent with Clay acting as a corrective context prior rather than a scar detector.}
  \label{fig:gradcam}
\end{figure*}

\textbf{Grad-CAM Interpretability:} To probe contribution at the fusion stage, we compute signed Grad-CAM at the CNN bottleneck (pre-fusion) and at the fusion output, and examine their difference $\Delta= $ fusion$-$pre-fusion (Fig.~\ref{fig:gradcam}). Across test chips, fusion adds positive contribution both inside (mean $\Delta$ = +0.049, $\sigma$ = 0.200) and outside (mean $\Delta$ = +0.100, $\sigma$ = 0.132) scar contours, suggesting a systematic, dataset-level tendency for fusion to broaden contribution outside scar interiors rather than concentrating it on them. This is consistent with Table~\ref{tab:ablation}: the hybrid architecture improves precision by approximately 9~pp (52.5\% to 62.2 $\pm$ 3.0\%) while only slightly reducing recall (69.9\% to 67.1 $\pm$ 0.6\%), indicating Clay's role is corrective rather than scar-detecting. Clay's broader semantic context suppresses false positives in landslide-similar background, while the CNN decoder handles pixel-level localization \cite{CTFNet,DGCFNet}.

\section{Discussion}
\label{sec:discussion}

The primary finding of this study is that Clay's pretrained representations improve landslide segmentation when
used as auxiliary context within a spatially-adept CNN backbone, but not when used as a standalone encoder. This
distinction has practical implications for how GFMs should be deployed in geohazard mapping applications. Architecture~2's test F1 of 64.5$\pm$1.8\% across $n=3$ seeds represents a meaningful advance over both the U-Net baseline (59.9\%, +4.6\,pp) and the Prithvi-EO-2.0 result of 60.7\% reported by Szwarcman \textit{et al.}\ \cite{ref6} on the same benchmark. This comparison is notable because Prithvi-EO-2.0 is currently the only other GFM evaluated on L4S, and Clay's hybrid integration outperforms it despite using a simpler fusion mechanism and roughly half the parameter count (339M versus 600M). Critically, the first-place competition result of 74.54\%
was achieved through test-set-informed preprocessing standardization and considerably more complex ensemble
architectures \cite{ref30}. Within the constraints of a clean train/validation/test protocol, Architecture~2's results are
competitive with published performance on this benchmark. Clay as a standalone encoder consistently underperforming the U-Net regardless of fine-tuning strategy
confirms that ViT global attention, while powerful for spectral representation, cannot substitute for the local
spatial priors that CNNs provide through hierarchical downsampling and skip connections.

Grad-CAM demonstrates physically interpretable attention. Spectral content of landslide instances is inherently ambiguous, since the spectral properties of bare soil overlap with those of plowed fields, riverbed sediments, and urban surfaces, producing systematic false positives in spectrally-driven detection \cite{brazil,multifusion}. Grad-CAM demonstrates this through how Clay appears to act as a geomorphic prior. This is consistent with suppressing detections in regions that are spectrally landslide-like but geomorphically implausible (e.g., flat agricultural fields or riverbeds). That terrain features contributed meaningful improvement despite low pixel-level discriminability suggests
the model learned to use geomorphic context to rule out false positives in spectrally similar flat
bare soil, a role the residual $\alpha$ (initialized to zero) allowed the model to discover independently. MC Dropout uncertainty concentrating at scar boundaries and transitional zones is physically
meaningful: crown and runout zones produce gradual spectral mixing that makes precise delineation ambiguous \cite{ref17}. These findings support a broader takeaway: GFMs provide spectral
generalization while CNNs provide spatial precision, and their combination is stronger than either
alone \cite{ref20,ref21}.

Several limitations constrain the present study. First, constant time and geolocation encodings are used for all
Clay inputs, meaning the model relies entirely on spectral information and cannot leverage geographic or seasonal
priors that Clay was pretrained to exploit. Incorporating real acquisition metadata could improve performance,
particularly for events with strong seasonal spectral signatures. Second, the classification threshold is selected by
manual post-training sweep on a small validation set (245 chips), introducing variance in the reported threshold
and potentially unstable optimal values. Third, despite Dirichlet calibration, Model~7a (mean $\sigma{=}0.002$) remains small, which could signal overconfidence, limiting the practical utility of MC Dropout uncertainty for flagging
unreliable predictions. Finally, the $128{\times}128$ chip size constrains spatial context, likely
contributing to poor performance on small isolated landslide patches.

Several directions could extend this work. Upsampling to $256{\times}256$ or using a sliding window approach on larger tiles could provide the broader spatial context necessary to improve localization; notably, the first-place competition model operated on
larger input resolutions \cite{ref30}. Pseudo-label training on unlabeled imagery from high-risk
regions could directly address the data scarcity that motivates this work. Deep ensembles would yield
more reliable uncertainty estimates and improve F1 through prediction averaging \cite{ref30}. Finally, cross-attention between Clay's spectral features and the TerrainPyramid could enable dynamic attention to specific geomorphic controls during decoding, replacing the static residual fusion used in both architectures.

\section{Conclusion}
\label{sec:conclusion}

This study evaluated whether Clay v1.5, a GFM, can improve pixel-level landslide segmentation on the
Landslide4Sense benchmark. The central finding is that foundation models are most effective when they complement
rather than replace spatially-detailed convolutional architectures. Clay as a standalone encoder (F1$=$55.2 $\pm$ 3.6\%)
underperformed the U-Net baseline (F1$=$59.9\%), while injecting Clay's pretrained representations as auxiliary
bottleneck context within a U-Net backbone achieved the best result of 64.5$\pm$1.8\%, competitive with current
state-of-the-art on this benchmark.

Two-stage training was essential for preventing destructive interference between encoder adaptation and decoder
learning. The Lov\'{a}sz--BCE loss and physically-informed terrain inputs both proved critical---the latter contributing
meaningful geomorphic context consistent with the physical hierarchy of
landslide initiation.

Grad-CAM analysis confirmed that the model learned representations physically consistent with how landslide scars
manifest spectrally and spatially, while MC Dropout revealed that uncertainty correctly tracks error-prone regions
despite overall overconfidence. Together these findings support a broader principle: in data-scarce Earth science
settings, GFMs provide spectral generalization while CNNs provide spatial precision, and their combination is
stronger than standalone structures. This hybrid principle likely extends to other geohazard and planetary surface mapping
tasks where spectral and topographic data carry complementary information, such as debris flow mapping or
volcanic deposit delineation using remote sensors.
The code used in this study is publicly available at \url{https://github.com/binhhuongvu/gfm-landslide-segmentation}.

\section*{Acknowledgments}
The author would like to thank Prof.\ S.\ M.\ Mousavi for his guidance and support throughout this research.

\begin{IEEEbiography}[{\includegraphics[width=1in,height=1.25in,clip,keepaspectratio]{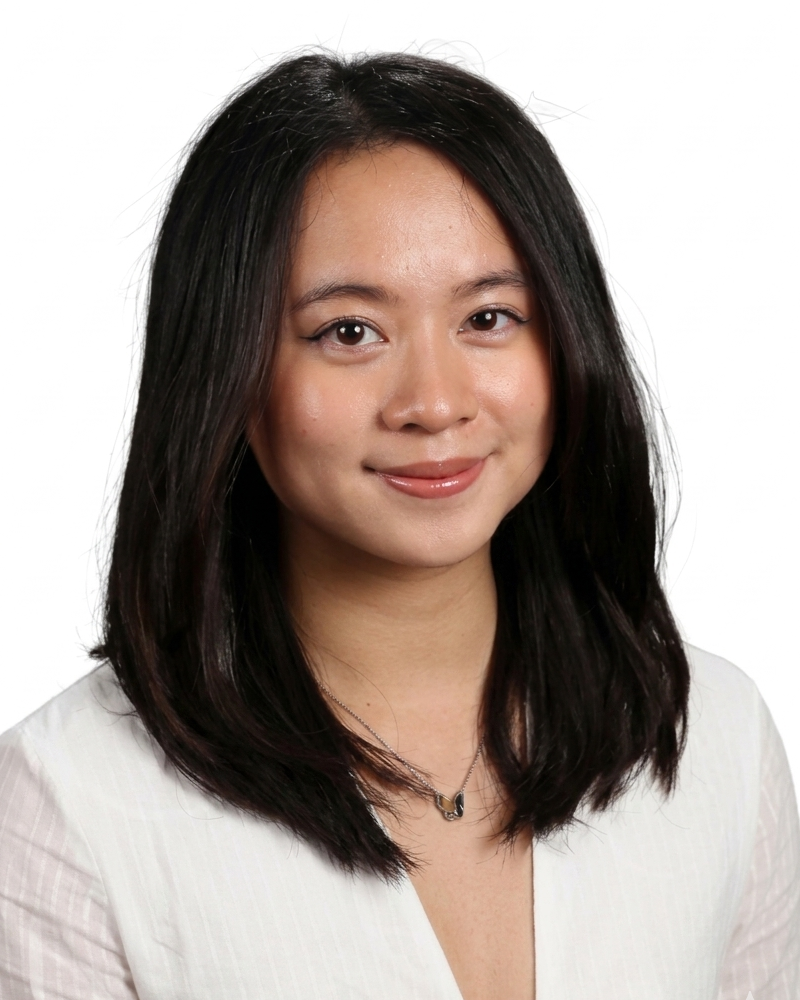}}]{Binh Huong Vu}
received a B.A. in Economics with a minor in Computer Science from Harvard University. Her research interests lie at the intersection of machine learning and humanitarian applications, with a focus on leveraging remote sensing for natural disaster prediction, management, and response in developing countries.
\end{IEEEbiography}
\vfill
\end{document}